# Triple Attention Mixed Link Network for Single Image Super Resolution

**Xi Cheng, Xiang Li, Jian Yang***
School of Computer and Engineering, Nanjing University of Science and Technology
{chengx, xiang.li.implus, csjyang}@njust.edu.cn

**Abstract**

Single image super resolution is of great importance as a low-level computer vision task. Recent approaches with deep convolutional neural networks have achieved impressive performance. However, existing architectures have limitations due to the less sophisticated structure along with less strong representational power. In this work, to significantly enhance the feature representation, we proposed Triple Attention mixed link Network (TAN) which consists of 1) three different aspects (i.e., kernel, spatial and channel) of attention mechanisms and 2) fusion of both powerful residual and dense connections (i.e., mixed link). Specifically, the network with multi kernel learns multi hierarchical representations under different receptive fields. The output features are recalibrated by the effective kernel and channel attentions and feed into next layer partly residual and partly dense, which filters the information and enable the network to learn more powerful representations. The features finally pass through the spatial attention in the reconstruction network which generates a fusion of local and global information, let the network restore more details and improves the quality of reconstructed images. Thanks to the diverse feature recalibrations and the advanced information flow topology, our proposed model is strong enough to perform against the state-of-the-art methods on the benchmark evaluations.

## Introduction

Single image Super-Resolution (SISR) is an important low-level computer vision task which have high practical value in many fields such as industrial inspection, medical imaging and security monitoring. SISR aims at recovering high-resolution image from only one low-resolution image. For this ill-posed inverse problem, widely used interpolation methods could not achieve visual pleasing results and many learning-based methods (Yang et al. 2010; Timofte et al. 2014) have been proposed. In the recent years deep-learning based algorithms (Dong et al. 2016) have been developed which greatly improved the super resolution quality and the details of the images could be better preserved with these powerful deep networks.

The introduction of the attention mechanism further improves the performance of the neural networks. SENet (Hu et al. 2017) and its derived super-resolution method (Cheng et al. 2018) focus on the attention between channels, and have achieved good results in many tasks. Attention is not limited to channels, the concurrent spatial attention and channel attention (Roy et al. 2018) have achieved better results in image semantic segmentation tasks. However, we found that in the super-resolution task, adding local attention like Roy et al. did not improve the image reconstruction performance, or even decrease the quality. Therefore, it is very important to seek a spatial or other attention that works effectively for super-resolution tasks. In order to recover more image details with the super-resolution models, global residual learning is widely used (Kim et al. 2016), while global residuals represent the image details predicted by the neural networks. Therefore, in order to help the network gain enhanced the details, we propose a spatial attention on the global residual. Also, special attention module works for multi parallel convolution kernel was also introduced. With these attention mechanisms, the performance of the model is significantly improved.

In recent studies, residual networks with deeper structure or networks with dense connections were used in image super-resolution (Tai et al. 2017; Tong et al. 2017; Zhang et al. 2018), both of which have achieved good results. Li et al. (Wang et al. 2018) found the commonality of the residual networks and dense networks, and proposed a Mixed Link structure that takes both of them into consideration, which improves the performance and reduces the number of model parameters. We introduced the attention enhanced mixed link block (AE-MLB) in our study. Unlike ordinary Mixed Link networks, our proposed structure was designed with appropriate zero padding to control the size of outputs equally among layers and we add attention to the channels between the connections of each layer. By introducing feature recalibration between channels, it enhances useful channel information and suppresses useless information. In addition, each logical layer contains two convolutional layers with different kernel sizes to gain different receptive field, and fuse kernel attention was introduced which can mix different output between convolutional layers to further improve the performance of the network. We also introduced multi supervise when training the network. Each AE-MLB will output high-resolution images and calculate the loss between output and target, so that our model could stably output high-resolution image with high quality. We summarize our contributions in the following points:

- We proposed a novel triple attention mixed link network (TAN) model for single image super resolution.

* Corresponding Author
Code will be upload to GitHub.

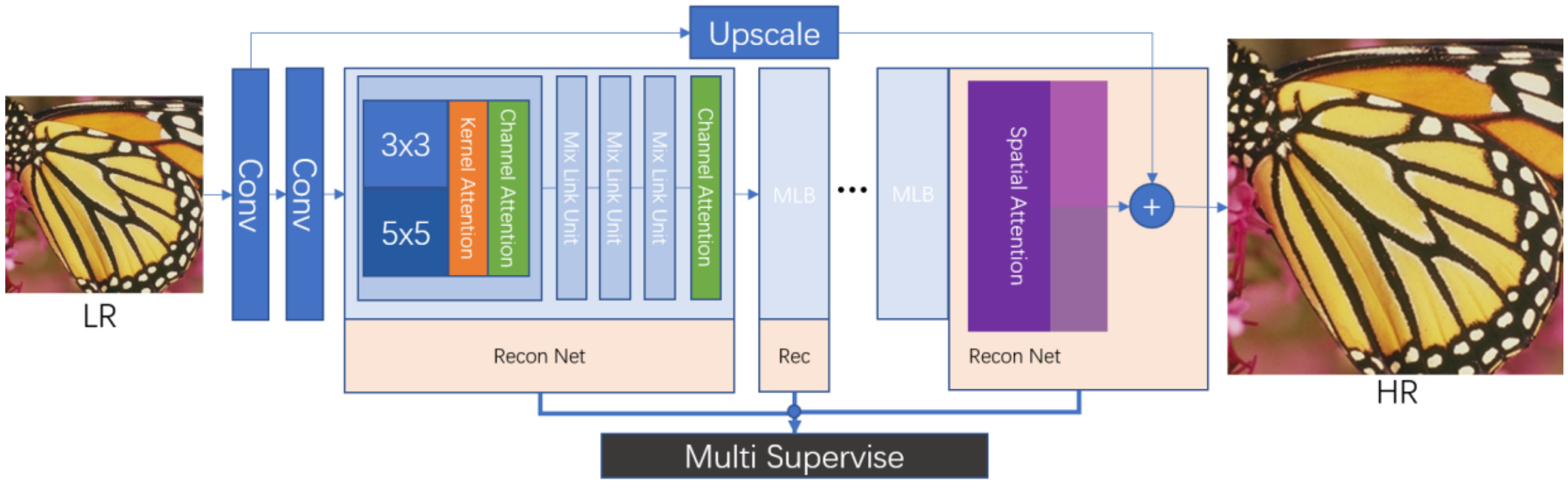

Fig.1 Overall structure of triple attention mixed link network

The global spatial attention (SA) and fuse kernel attention (KA) we proposed could significantly improve the super resolution performance.

- We proposed the attention enhanced mixed link block which could help achieve better performance with less model parameter.
- Our model achieved state of the art performance according to serval benchmark datasets.

## Related Works

Single image super resolution is a research hot spot in recent years. Deep-learning based methods showed great improvement compared with conventional methods such as interpolations, anchored neighborhood regression (Timofte et al. 2014), self-exemplars (Huang et al. 2015) and methods based on sparse encoding (Yang et al. 2010).

SRCNN (Dong et al. 2016) firstly used convolutional neural networks to sample images and achieved significant improvements. The performance of SRCNN was limited by its shallow structure. To achieve higher performance the networks are tend to be deeper and deeper, Kim et al. proposed the VDSR (Kim et al. 2016) model with a deeper structure. In order to make the deep model trainable, recursive supervision and residual models were introduced (Kim et al. 2016; Tai, Yang and Liu. 2017). In recent years, some very deep models have been proposed such as EDSR and MDSR (Lim et al. 2017), which achieves very pleasing performance on super-resolution tasks. In addition, super-resolution models integrated with Dense connections have been proposed, such as SRDenseNet (Tong et al. 2017) and MemNet (Tai et al. 2017), which can effectively utilize different levels of features. Later, the Residual Dense Network (RDN) (Zhang et al. 2018) was proposed which makes more use of hierarchical features, in addition, the structure can also effectively control parameter growth and makes large-scale models trainable. In terms of reconstruction network, the model also gradually uses deconvolution and effective subpixel shuffle (Shi et al. 2017; Lai et al. 2017) to replace the traditional pre-interpolation process, which simplifies the computational complexity and further improves the performance of the model.

The above methods showed impressive super resolution performance, however their structures were complex and very deep. To achieve higher performance, attention is another key role other than the scale and complexity of network structure. Moreover, the above model is not sufficient for the use of hierarchical information. Feature extraction for an LR image often requires different receptive fields and effective recalibration and fusion of features extracted from different subfields, which was often overlooked in previous super-resolution models. In order to solve these problems, we proposed a triple attention network with mixed link structure for single image super resolution task and we will detailly introduce our proposed TAN in the next section.

## Proposed Method

### Overall model structure

As shown in Fig.1, our proposed triple attention mixed link network contains three basic parts, which are shallow feature extractor (SFENet), attention enhanced mixed link blocks (AE-MLBs) and reconstruction networks with multi supervise. The SFENet contain two convolution layers to grab the shallow features though the network. Low resolu-tion images fed directly into the network and divided to two branches, one was input to the upscale module after the first convolution layer in SFENet and the other was then pass through the second convolution layer and input to the AE-MLBs for predict the details. The reconstruction network combines the upscaled image with the predicted details to generate the high-resolution image.

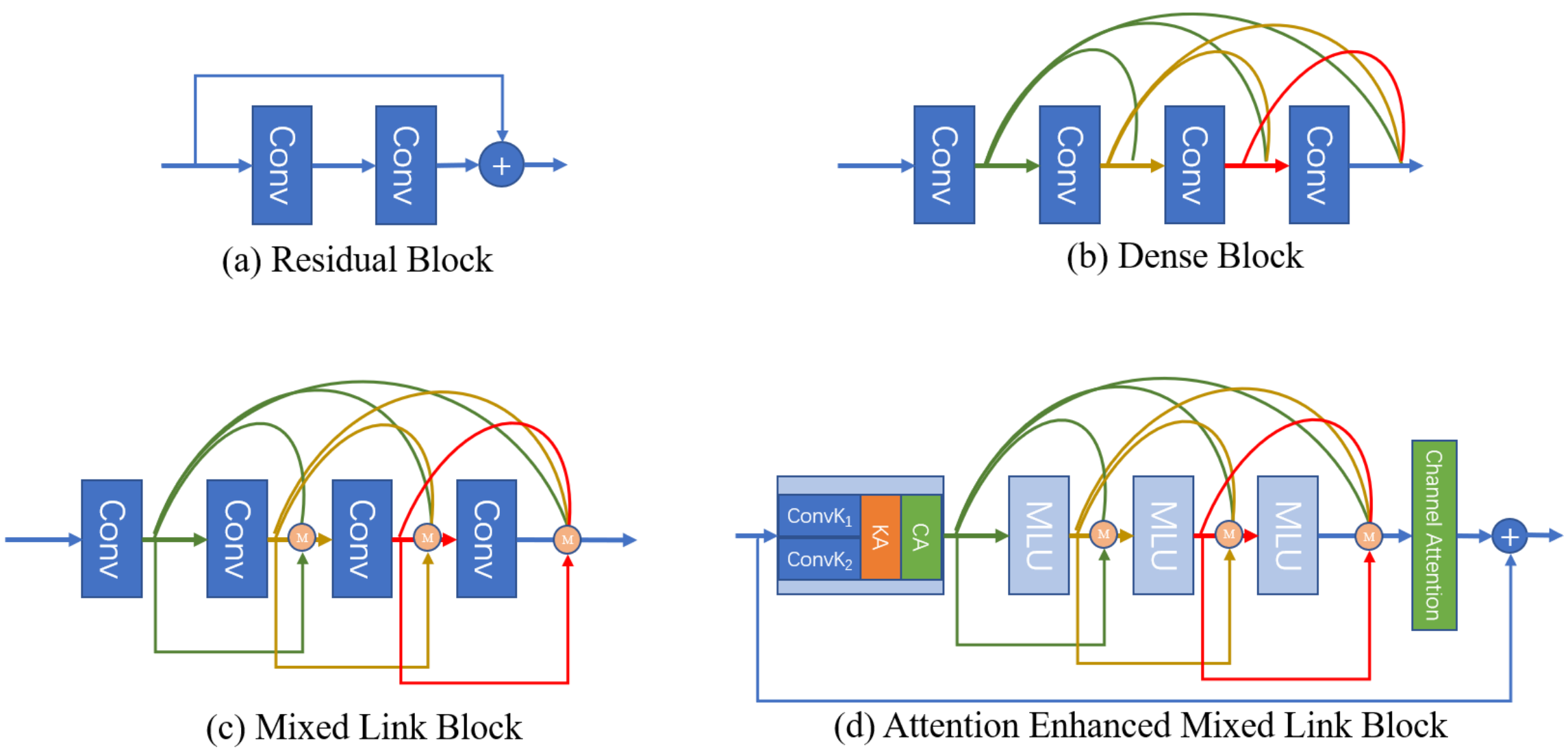

Fig. 2 Comparison of different network structures

## Mixed Link Connections

As shown in Fig.1 (d), operator M denotes the mixed link connection which could be calculated as formula1-4.
This operation could be divided into three parts. The first one slice the input channels to two parts equally, $S(.)$ means the slice operation in formula 1.

$$x_i^1, x_i^2 = S(x_i) \quad (1)$$

Second, the output of one layer or unit should also be sliced into two equal parts in the channel dimension. In Fig. 2, structure (c) and structure (d) could be calculated as formula 2 and 3.

$$\hat{x}_i^1,\ \hat{x}_i^2 = S\big(\sigma W_1(x_i)\big) \quad (2)$$

$$\hat{x}_i^1,\ \hat{x}_i^2 = S(A_c(A_k(\sigma W_1(x_i) + \sigma W_2(x_i)))) \quad (3)$$

Where $A_c$ and $A_k$ denotes the channel and kernel attention, $\sigma$ means the PReLU [18] activation function. $W_1$ and $W_2$ means convolution layers with 3x3 and 5x5 kernel.

$$x_{i+1} = C\left(x_i^1, C\Big((\hat{x}_i^1 + x_i^2),\ \hat{x}_i^2\Big)\right) \quad (4)$$

The final step is shown as formula 4. $C(.)$ denotes the concatenate operation and this enable the network to be partial residual network and partial dense network.

## Triple Attention

### Channel attention

Channel attention could help the network gain the ability of modeling and selecting the information among channels which is also called feature recalibration. This was proved to be effective to improve the performance of the model on the field of image recognition and restoration. As shown in Fig.3 (a), the channel attention module consists with one global average pooling layer which squeeze the features spatially to grab the global information among channels, then two 1x1 convolution layers named ConvD and ConvU generate a bottleneck. Finally, a Sigmoid activation layer to map the information between 0 and 1 and the output is used to reweight the original output to generate a self-learned channel wise attention. The process of channel attention could be calculated as the following formulas:

$$S(x_c) = \frac{1}{HW}\sum_{i}^{H}\sum_{j}^{W} x_c(i,j) \quad (5)$$

Where $S(.)$ represents the spatial squeeze with global average pooling. H and W denotes the height and width of the feature map. $x_c$ means channel c of the input feature map $x$.

$$A_c(x) = \sigma^s\big(W_u\sigma^p(W_d S(x))\big) * x \quad (6)$$

Where $A_c(.)$ denotes the channel attention, $\sigma^s$ means the sigmoid activation function and $\sigma^p$ means the PReLU activation function (He et al. 2015), $W_u$ and $W_d$ means the two convolution layers with 1x1 kernel size.

### Kernel Attention

Different sizes of convolution kernels can provide different receptive fields, extracting different features. Many networks utilized different size of convolution kernels to improve the performance (Szegedy et al. 2017). Therefore, in order to improve the super-resolution capability, we use 3x3 and 5x5 convolution kernels, and use 1 and 2 zero padding to ensure that the feature map size of each layer is equal. In addition, we use kernel attention to perform feature recalibration on the channels output from layers with different kernel. Kernel attention is actually a special channel wise attention, the structure could be seen as Fig.3 (b). The operation process can be derived from formula 7 and formula 8.

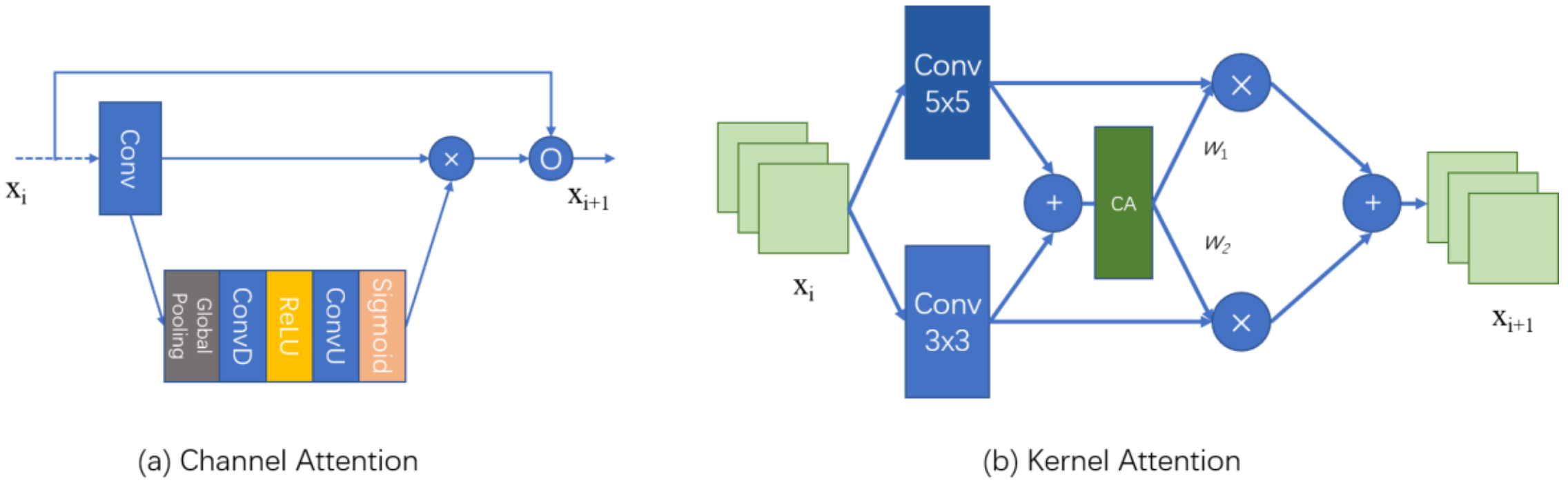

Fig.3 Channel attention and Fuse kernel attention, 'O' denotes add or concatenate operation

$$W_k = A_c(\sigma W_1(x_i) + \sigma W_2(x_i)) \quad (7)$$

$$A_k(x_i) = W_k * (\sigma W_1(x_i) + \sigma W_2(x_i)) \quad (8)$$

**Global spatial attention**

In addition to channel attention, the spatial attention model proved to be effective in the segmentation task (Roy et al. 2018). However, we found that the performance of the super-resolution task does not improve. We have tried a variety of solutions that incorporate spatial attention. However, we all found a slight drop in model performance. We blame this phenomenon on the local spatial attention only focus on local information, and cannot play the role of filtering effective global information. To solve this problem, a fusion for global information and local information is needed.

The reconstruction part of the super-resolution model in this paper adopts the strategy of global residual learning. The global residual channel is increased to 2 times of the original. The half of the channels are weighted by Global information, and the other half retains the local information. Then the two are summed and averaged to achieve global and local information fusion. This process could be seen as Fig.4. Finally, the subpixel shuffle is used for up sampling, and each block is weighted and fused.

$$R_1, R_2 = S(M(x)) \quad (9)$$

$$\mathrm{HR} = \sum_{b}^{B} w^b \frac{1}{2}\left(R_1 + R_2 G(M(x))\right) + U(I_L) \quad (10)$$

**Multi Supervise and Loss Function**

Multi supervise are used during training process. For each AE-MLB, a high-resolution image is generated and the loss is calculated. Finally, the loss values of all the blocks are added and the arithmetic mean is calculated as the overall loss. This could be calculated as formula.11.

$$Loss = \frac{1}{NB}\sum_{i=1}^{N}\sum_{b=1}^{B} p\left(\hat{y}_b^i - y_b^i\right) \quad (11)$$

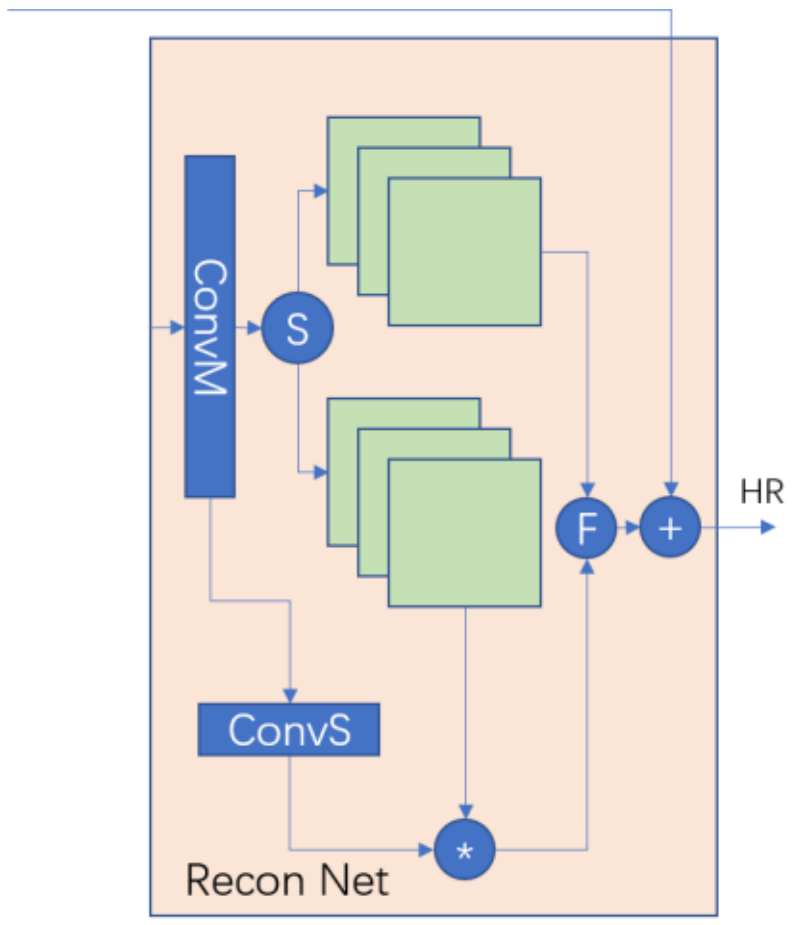

Fig.4 Reconstruction net with spatial attention

Where N means the batch size, B means the number of AE-MLBs, $\hat{y}_b^i$ and $y_b^i$ means the ground truth and the high-resolution image generated from the network. $p(\mathrm{x})$ is the Charbonnier penalty function, $\varepsilon$ is set to 0.001 and this can be calculated as the following formula:

$$p(\mathrm{x}) = \sqrt{x^2 - \varepsilon^2} \quad (12)$$

## Experiment

**Datasets and Training Details**

We used DIV2K (Agustsson et al. 2017) dataset for training the network, which contains 800 high resolution images and we used Set5 (Bevilacqua et al. 2012), Set14 (Zeyde et al. 2010), BSD100 (Martin et al. 2001), Urban100 (Huang et al. 2015) and Manga109 (Matsui et al. 2017) for testing. All RGB images were converted to YCbCr color space and we selected the Y channel for training and testing the super resolution model. For higher scale super resolution, we first train the model with lower factor and then finetuning the model with higher factor with the pretrained checkpoint. We used PyTorch 0.4.0 as the deep-learning framework to build the neural network and we used a server with NVIDIA Tesla P40 GPU as the training setup. Also, parallel method was utilized. We equally divided the training batch to 5 GPUs and this greatly accelerate the model training process. In this work, we perform self-ensemble (Lim et al. 2017) to gain higher performance. We flipped and rotate the test images to augment 7 images from the original. We input these images into the network and performed inverse transform on the output high resolution images. All these images were then added together and averaged to get the final high-resolution output.

Table.1 Compare with state-of-the-art methods

| Dataset | Scale | Bicubic PSNR/SSIM | A+ PSNR/SSIM | SRCNN PSNR/SSIM | LapSRN PSNR/SSIM | MemNet PSNR/SSIM | EDSR PSNR/SSIM | RDN PSNR/SSIM | TAN (Ours) PSNR/SSIM |
|---|---|---|---|---|---|---|---|---|---|
| Set5 | x2 | 33.65/ 0.931 | 36.54/ 0.955 | 36.65/ 0.955 | 37.52/ 0.959 | 37.78/ 0.960 | 38.11/ 0.960 | **38.24/ 0.962** | **38.31/ 0.965** |
| | x4 | 28.43/ 0.811 | 30.32/ 0.860 | 30.50/ 0.863 | 31.54/ 0.885 | 31.74/ 0.889 | **32.46/ 0.897** | **32.47/ 0.899** | 32.41/ 0.896 |
| Set14 | x2 | 30.34/ 0.870 | 32.40/ 0.906 | 32.29/ 0.908 | 33.08/ 0.913 | 33.28/ 0.914 | 33.92/ 0.920 | **34.01/ 0.921** | **34.07/ 0.925** |
| | x4 | 26.01/ 0.704 | 27.34/ 0.751 | 27.52/ 0.753 | 28.19/ 0.772 | 28.26/ 0.772 | 28.80/ 0.786 | **28.81/ 0.787** | **28.82/ 0.796** |
| BSD 100 | x2 | 29.56/ 0.844 | 31.22/ 0.887 | 31.36/ 0.887 | 31.80/ 0.895 | 32.08/ 0.898 | 32.32/ 0.901 | **32.34/ 0.901** | **32.41/ 0.907** |
| | x4 | 25.97/ 0.670 | 26.83/ 0.711 | 26.91/ 0.712 | 27.32/ 0.727 | 27.40/ 0.728 | **27.71/ 0.742** | **27.72/ 0.742** | 27.66/ 0.741 |
| Urban 100 | x2 | 26.88/ 0.841 | 29.23/ 0.895 | 29.23/ 0.895 | 30.41/ 0.910 | 31.31/ 0.920 | 32.93/ 0.935 | **33.09/ 0.937** | **33.13/ 0.939** |
| | x4 | 23.15/ 0.660 | 24.34/ 0.721 | 24.53/ 0.725 | 25.44/ 0.756 | 25.50/ 0.763 | **26.64/ 0.804** | 26.61/ 0.803 | **26.73/ 0.809** |
| Manga 109 | x2 | 30.86/ 0.936 | 35.37 / 0.968 | 35.82 / 0.969 | 37.27 / 0.974 | 37.72/ 0.974 | 39.10/ 0.977 | **39.18/ 0.978** | **39.59/ 0.981** |
| | x4 | 24.93/ 0.790 | 27.03 / 0.851 | 27.83 / 0.866 | 29.09 / 0.890 | 29.42/ 0.894 | **31.02/ 0.915** | 31.00/ 0.915 | **31.04/ 0.919** |

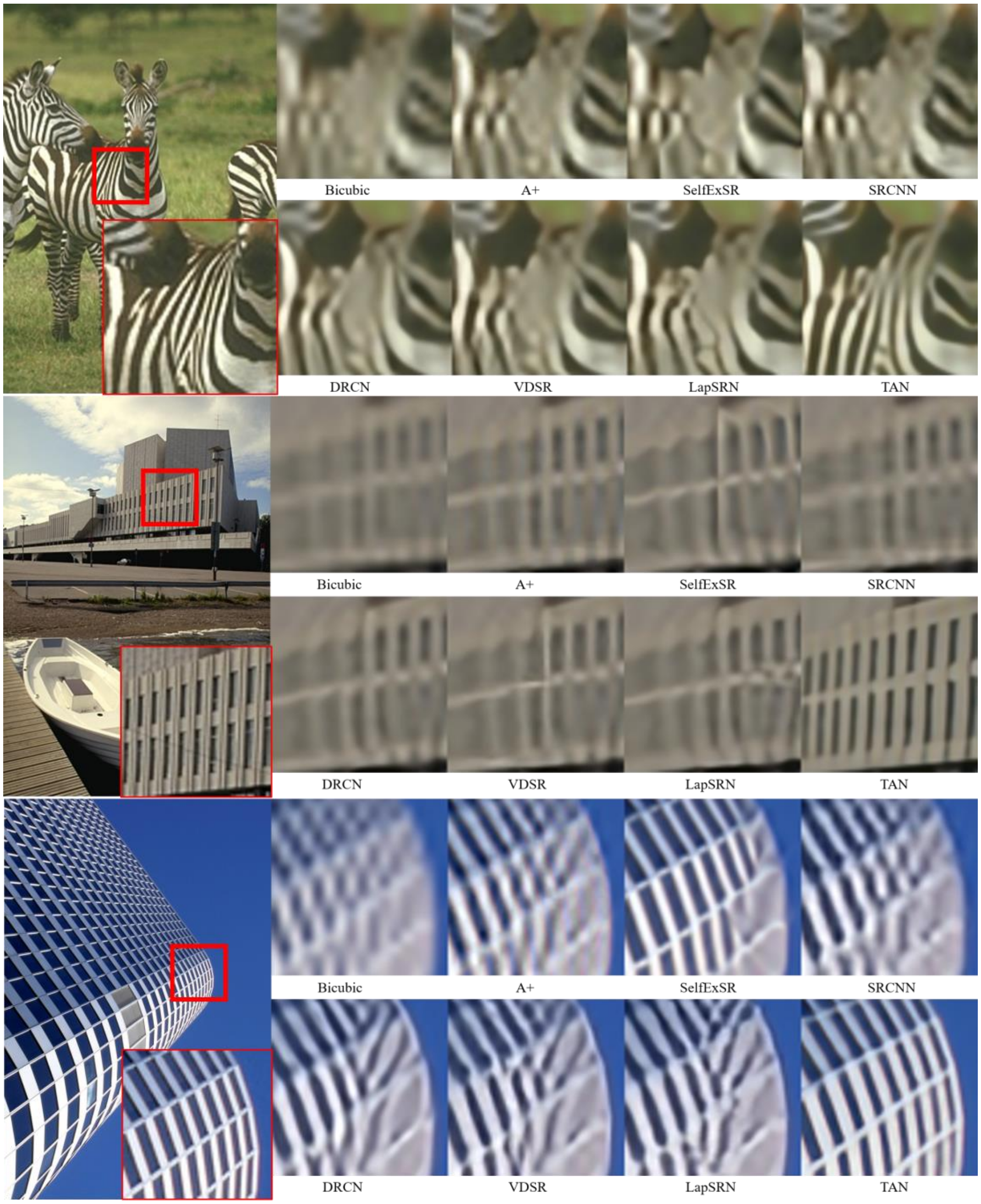

Fig.4 Visual comparison for details of reconstructed images

### Compare with state-of-the-art methods

We used peak signal to noise ratio (PSNR) and structural similarity (SSIM) (Wang et al. 2004) as the image quality metrics and we compared the result generated by our proposed TAN with Bicubic, A+ (Timofte et al. 2014) and deep learning based super resolution methods including SRCNN (Dong et al. 2016), LapSRN (Lai et al. 2017), MemNet (Tai et al. 2017), EDSR (Lim et al. 2017) and RDN (Zhang et al. 2018). Table1 show the quantitative results on Set5, Set14, BSD100 and Urban100 with scale factor x2 and x4. The best was marked with red and the second was marked with blue. Our proposed method could perform against other methods among the 5 datasets in scale x2 task and also achieved favorable results in scale x4 task.

We also illustrate a visual quality comparison to show the reconstructed details on BSD100 and Urban100 datasets. As shown in Fig.4, the red rectangle represents where the sub image was taken from and the ground truth is located on the right bottom and marked with red edge. The first two images were img_052 and img_092 from BSD100 dataset and the third one is img_005 from Urban100 datasets. The textures of theses images were reconstructed clear with our proposed TAN while images generated from other methods were blurred, distorted and have error with the image details. This illustrates our proposed method could generate high resolution images with more accurate details.

### Model Parameter

We also compared the parameter of our proposed model with the state of the arts, the results of scale x2 super resolution on Set5 could be seen in Fig.5. The x axis shows the million parameters and the y axis shows the PSNR result. Our proposed method could achieve obvious higher results compared with those recursive model (MemNet (Tai et al. 2017)) or model with small quantity of convolutional layers (LapSRN (Lai et al. 2017)) and model with encoder-decoder framework (RED30 (Mao et al. 2016)). Also, our proposed TAN could perform against the state-of-the-art among the large-scale networks with only 7.2 million model parameters, which is relative light weight and is 83.3% less than EDSR (Lim et al. 2017) (43M), 67.3% RDN (Zhang et al. 2018) (22M) and 28% less than DBPN

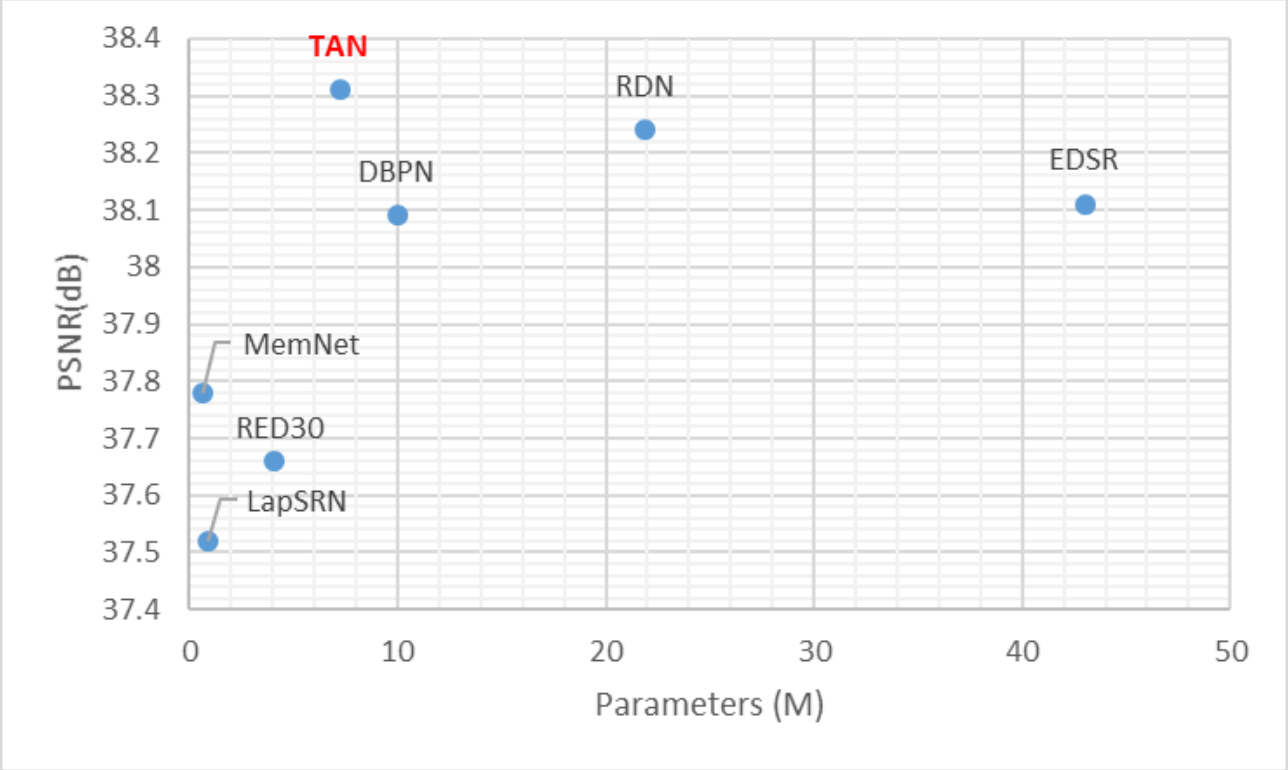

Fig.5 Comparsion of model parameters and performance

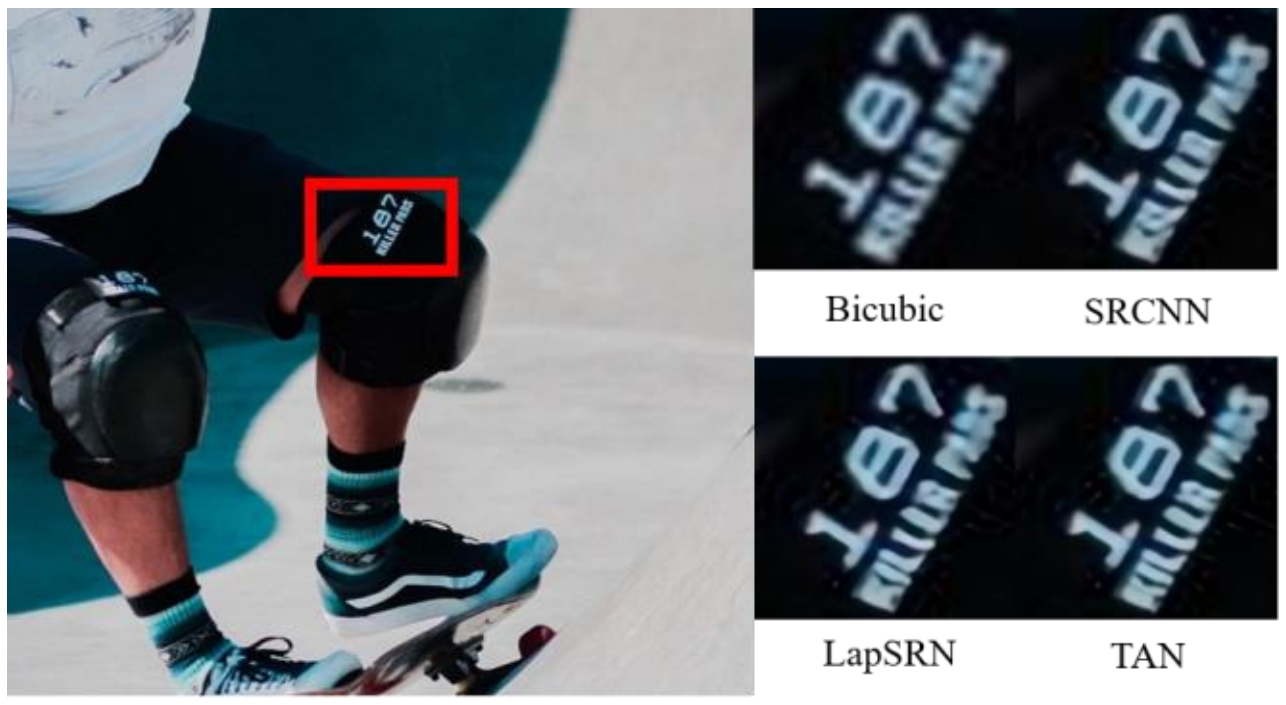

Fig.6 Visual comparison of real-world photo with scale factor x4

(Haris et al. 2018) with 10M network parameter.

### Super-resolving real-world images

In this section, we conduct experiment on image from the real-world. The image was compressed with JPEG method and with unknow degradation model and unknow high resolution ground truth. As shown in Fig.6, we compare our result with bicubic interpolation, SRCNN (Dong et al. 2016) and LapSRN (Lai et al. 2017) and our proposed TAN generates more clear and sharp details and edges among these methods.

## Ablation study

### Model structure

In this section we study the effect of network structure on the performance of super resolution models. We define the baseline model with 6 mixed link blocks and no attention modules were include. We replaced the mixed link connections with pure concatenate and skip layer connections to create a dense network and a residual network. We trained the models on DIV2k dataset and we compare the performance of these network structures and the result was shown in the Table.2. The experiment result shows that Mixed Link structure could achieve higher performance and less parameters.

Table.2 Model parameter and performance on Set5 with different network structures.

| Structure | Baseline (MixNet) | ResNet | DenseNet |
|---|---|---|---|
| Parameter | 1.86M | 2.03M (+0.17) | 2.55M (+0.69) |
| Set5 | 37.81 | 37.69 (-0.12) | 37.74 (-0.07) |
| Set14 | 33.40 | 33.28 (-0.12) | 33.32 (-0.08) |
| BSD100 | 32.05 | 31.96 (-0.09) | 31.98 (-0.07) |

Table.3 Performance with different attention modules

| Module | Set5 Scale x2 | Set14 Scale x2 | BSD100 Scale x2 |
|---|---|---|---|
| Baseline | 37.81 | 33.40 | 32.05 |
| CA | 37.89 (+0.08) | 33.45 (+0.05) | 32.08 (+0.03) |
| KA | 37.95 (+0.14) | 33.54 (+0.14) | 32.14 (+0.09) |
| SA | 37.91 (+0.10) | 33.49 (+0.09) | 32.12 (+0.07) |
| CA+KA | 38.19 (+0.38) | 33.82 (+0.42) | 32.27 (+0.22) |
| CA+SA | 38.15 (+0.34) | 33.54 (+0.14) | 32.17 (+0.12) |
| KA+SA | 38.09 (+0.28) | 33.66 (+0.26) | 32.22 (+0.17) |
| CA+KA+SA | 38.30 (+0.49) | 33.97 (+0.57) | 32.37 (+0.32) |

Table.4 Performance with different number of blocks

| Number Blocks | Set5 Scale x2 | Set14 Scale x2 | BSD100 Scale x2 |
|---|---|---|---|
| 1 | 37.97 | 33.53 | 32.15 |
| 2 | 38.12 | 33.69 | 32.26 |
| 3 | 38.23 | 33.79 | 32.32 |
| 4 | 38.28 | 33.87 | 32.34 |
| 5 | 38.29 | 33.96 | 32.34 |
| 6 | 38.30 | 33.97 | 32.37 |

### Attention module

In this section we study the attention modules we introduced in our network. We researched on the effects of these attention modules. We first train models with only one type of attention module and then different combination of these modules. There are six different combinations and the results for scale factor x2 with three test datasets are shown in Table. 4. The first line shows the PSNR score and the second line shows the improvement. Baseline means a pure mixed link network with no attention module, CA denotes the channel attention, KA means the fuse kernel attention, SA means the global spatial attention. There is an obvious increase of performance when adding these attention modules to the baseline network among the experiment results of three testing datasets. When using one attention module, there will be at most 0.14dB gain in Set14, 0.42dB with two module and 0.57dB with all of the three attention modules.

### Study the number of AE-MLBs

The number of attention-enhanced mixed link blocks could directly affect the scale and layers of the network and determines the model parameter and the super resolution performance. We studied on the number of blocks and the PSNR for scale x2 on Set5, Set14 and BSD100 is shown in Table.4. The experiment result shows the performance is better with more blocks. Our proposed TAN allows to train deeper network and which enable our model to grab more information from the images and predict more accurate details to reconstruct high resolution image with favorable quality.

## Conclusion

In this work, we propose a novel single image super resolution method which utilized mixed link connections and three different attention including channel attention, fuse kernel attention and global spatial attention and we name the model triple attention mixed link network (TAN). The mixed link structure helps the network gain stronger representation ability and proved to be more powerful than residual or dense networks. Moreover, the attention mechanisms give an impressive improvement on performance. The channel attention (CA) could recalibrate the information among channels. The fuse kernel attention (KA) could fuse the feature output from layers with different kernels, which enable the model gain different receptive field. The global spatial attention mixes the information from local parts and global parts of channels which greatly improved the reconstruction network. With these attention modules, the proposed attention enhanced mixed link blocks (AE-MLB) was utilized as the basic part to build the whole network. Thanks to the sophisticated network structure and effective attention mechanisms, our model could perform against the state of the arts according to serval benchmark evaluations.

## Acknowledgement

This work was supported by the National Science Fund of China under Grant Nos. U1713208 and 61472187, the 973 Program No. 2014CB349303, and Program for Changjiang Scholars.